# Hilditch's Algorithm Based Tamil Character Recognition

V. Karthikeyan
Department of ECE, SVS College of Engineering
Coimbatore, India,
Karthick77keyan@gmail.com

**Abstract**-Character identification plays a vital role in the contemporary world of Image processing. It can solve many composite problems and makes human's work easier. An instance is Handwritten Character detection. Handwritten recognition is not a novel expertise, but it has not gained community notice until Now. The eventual aim of designing Handwritten Character recognition structure with an accurateness rate of 100% is pretty illusionary. Tamil Handwritten Character recognition system uses the Neural Networks to distinguish them. Neural Network and structural characteristics are used to instruct and recognize written characters. After training and testing the exactness rate reached 99%. This correctness rate is extremely high. In this paper we are exploring image processing through the Hilditch algorithm foundation and structural characteristics of a character in the image. And we recognized some character of the Tamil language, and we are trying to identify all the character of Tamil In our future works.

**Keywords-** Character Identification, Hilditch Algorithm, Hand written Recognition, Contemporary

## I.     INTRODUCTION

Visualize an inhabitant walking into a rustic Internet booth, who may be semi-literate or even illiterate, wanting to use the power of the Internet to either converse with a comparative somewhere else, or get in touch with a city hospital, or get vital crop information. At present it is not obtainable English-based keyboard and applications are entirely unknown and unapproachable. As a result he or she feels push to out and not part of the ongoing in sequence revolution. On the other hand, had the Computer been able to believe applications in the local language, this typical villager would have been as contented using it as anyone else. Computers have become an essential part of many facets of our lives. However, in the Indian context the use of computers is far less compared with that in the developed nations of the West because of the reason we have already hinted at: the language of the interface is almost always English and the communication is in the "written" form, i.e., via the keyboard. Barely 65 % of our population is literate, of which only an elite minority (~5%) can read, write, and speak the English language. This shuts out most of the Indian population from the World Wide Web and its huge potential. The current keyboard has been developed for English and cannot be naturally adapted for Indian languages. Hence there is a need for a handwriting interface. In the West, the handwriting interfaces is already available for English, these are for some specific applications. Whereas, in the Indian context, these interfaces must be part of main stream applications. In this paper we are exploring image processing through the Hilditch algorithm basis and structural characteristics of a character in the image. And we identified some character of the Tamil language, and we are trying to identify all the character of Tamil.

## II.     IMAGE PREPROCESSING STAGE

The defaulting dimension of picture is in use into description to recognize the particular Tamil character. This phase has grim of steps and procedure. They are briefly explained below.

### A.  Binarisation

Binarisation is a method by which the grey scale images are transformed to binary images.  The most ordinary process is to choose an appropriate threshold for the image and then translate all the intensity morals above the threshold intensity to one concentration value representing either "black" or "white" value i.e. 0 or 1.

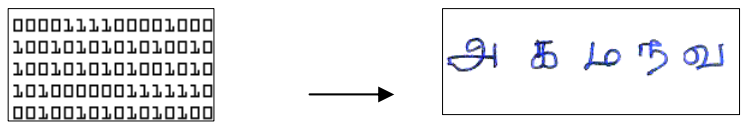

Fig.1. Binary values of the corresponding Tamil word

### B.  Segmentation

An input image consists of a consistent manuscript neighborhood with different text appearance. In the process, it is broken down into ingredient text lines, words finally into entity characters. This method is based on the horizontal projection outline of the image. Black values in the projection profile correspond to the parallel gaps among passage lines. Each text column is recognized using two location lines; the higher line and the inferior line. They match to the least value and greatest zero value positions neighboring a text line, respectively. (See Fig. 1) We extract two more reference lines from each text line, namely, the upper baseline and the lower baseline. For this, we use a method similar to [1]. First derivative of the horizontal projection profile is calculated for each segmented text line. The restricted intense of the first imitative in the two halves of the text line are taken to be the two baselines. The lines drawn across the two peaks indicate the two baselines. Words and characters are segmented using the perpendicular projection outline of each text line. Word borders and character borders are noticeable as the previous are much wider than the final. The algorithm is based on a particular attention to straightening out hand-printed characters which was "stirring" due to reduced scanner resolution. The program initially looked for characters, which were not stirring (separated by columns of zeros).

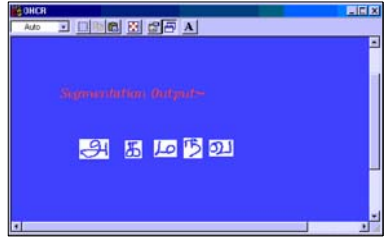

Fig.2. Line segmentation using horizontal and vertical projection outline

*C. Bounding Box*

Before examining the characters, it is significant to recognize the (pixel) boundaries of the particular character. Therefore, a bounding box has to be recognized for every character. For this, we primarily compute the row-wise parallel projection of the complete text file and classify begin and finish positions of every line, from the valleys in the projection. Having found the line pixels, we find the perpendicular projection of each line, the valleys of which show the boundaries of every character in the line [2]. While, each of the characters in a line might not be of identical height, so we consider only an estimated bounding rectangle for the characters. This is a critical stage in instant based identification methods and can be intended from a perpendicular projection of the bounding box.

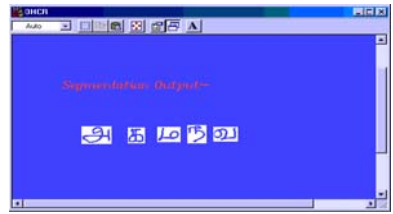

Fig.3. Bounding of each character

*D. Normalization*

After analyzing each character, it is significant to identify the boundaries of individual character. Thus, a bounding box is identified for each character. Similarly, each one of the character in a line might not be in the same height, what we have now processed is only an approximate bounding box for the characters. So we must make the each character at same height we go for Normalization. It is the process of giving accurate size and equal height to each character in the identified image.

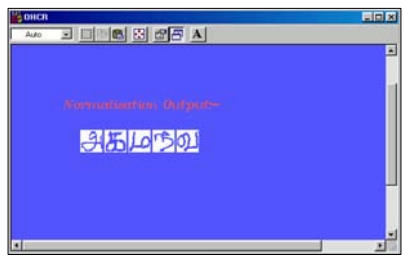

Fig.4. Normalized character Image

*E. Skeletonization*

Skeletonization is the method of detaching off of an outline as a lot of pixels as possible as lack of disturbing the common form of the outline. In other words, subsequent to pixels have been detached off, the outline should still be recognized. Hence we use the method for detaching the outline by the Skeletonization process is used to obtain as thin as possible, connected and centered when these are satisfied, then the algorithm must stop.

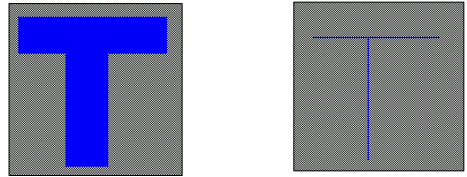

Fig.5. Image Skeletonization

### III. HILDITCH'S ALGORITHM

Skeletonization is helpful when we are paying attention not in the size of the outline but quite in the relative location of the strokes in the outline (*Character Recognition, X, Y chromosome Recognition*) there are several algorithms which were designed for the proposed work. In this projected work we are so familiar with the Hilditch's Algorithm. Consider the following 8-neighborhood of a pixel p1

|  P9 | P2 | P3 |
| --- | --- | --- |
|  P8 | P1 | P4 |
|  P7 | P6 | P5 |

Fig.6. 8-neighbourhood pixel Image

Using Hilditch's Algorithm it decides whether to peel off q1 or remain it as part of the resultant skeleton. For this reason we place the 8 neighbours of p1 in a clockwise order and the functions are defined bellow

A (p1) = number of non-zero neighbours of p1.

X(p1) = number of 0,1 patterns in the sequence p2, p3, p4, p5, p6, p7, p8, p9, p2.

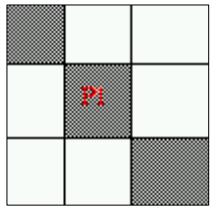 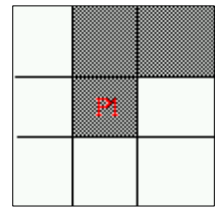

*A(p1)=2,A(p1)=1*                                              *A(p1)=2,A(p1)=2*

There are two versions for Hilditch's algorithm, one using a 4x4 window and the other one using a 3x3 window. Here we considered with the 3x3 window version. Hilditch's algorithm consists of performing numerous passes on the outline and on each pass; the algorithm checks all the pixels and decide to change a pixel from 0 to 1 if it satisfies the following four conditions:

- 2 < = A(p1) < = 6
- X (p1)=1
- p2.4p.8=0 or X(p2)!= 1
- . .p. .p6=0or ((p4)!=1

Stop when nothing changes (no more pixels can be removed)

**Condition1:** 2<=A (p1) <=6

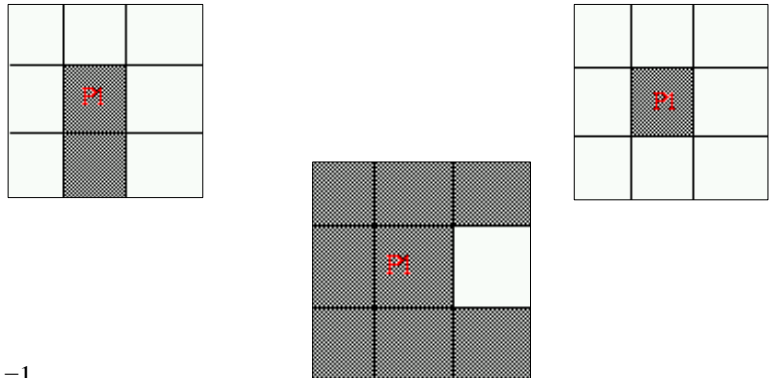

**Condition 2:** B (p1) =1

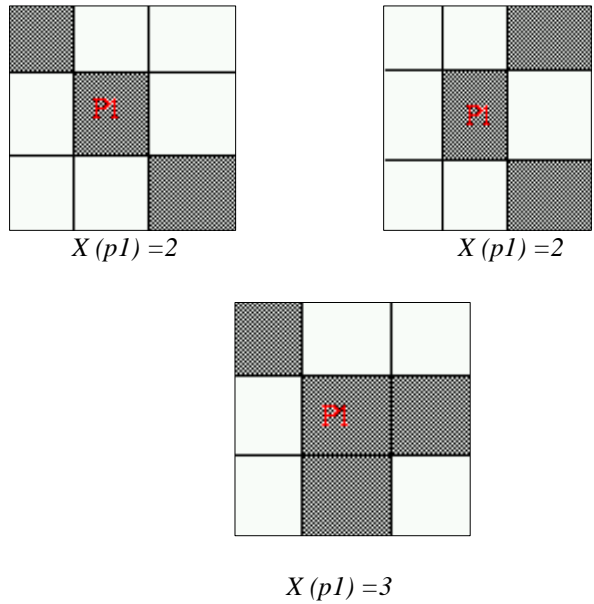

$X(p1) = 2$     $X(p1) = 2$

$X(p1) = 3$

**Condition 3:** p2.p4.p8 = 0 **or** X (p2)! =1

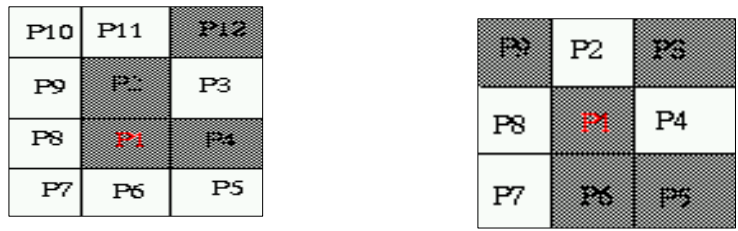

Here is one example where X (p2) is not 1 p2.p4.p8=0.
This condition ensures that 2-pixel wide vertical lines do not get completely eroded by the algorithm.

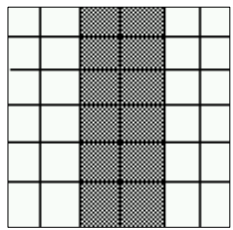

**Condition 4:** p2.p4.p6 = 0 **or** X (p4)! =1

Where X (p4)! =1.   Where p2.p4.p6=0.

## IV.  CHARACTER RECOGNITION

For a binary image with only two gray levels, 0 and 1, M00 gives the whole region of the image. The two first categorized arithmetical moments, M01 and M10, corresponds to the middle group of an image function f(x, y). The middle group is the position where all the group of the image could be determined without changing the first moment of the image about an axis. In terms of moment values, the coordinates of the middle of mass are given as X = M10 / M00 and Y = M01 / M00. Which define a distinct the location of the image that can be used as a reference point to describe its location. Referencing to the properties of lower order moments, a set of four non-linear combinations of lower order arithmetical moments is proposed here to represent all Tamil characters in the image level surface (x, y) as the recognition features. A point (f1, f2, f3, f4) in this four dimensional space represents one Tamil character

$$ƒ1 = M20 + M02 + M00 \qquad (1)$$
$$ƒ2 = (\sqrt{(M20 - M02)}\ \hat{\ }2) + M11 \qquad (2)$$
$$ƒ3 = \sqrt{(M10 - M01)}\ \hat{\ }2 \qquad (3)$$
$$ƒ4 = M30 + M03 \qquad (4)$$

## V.  CHARACTER RECOGNITION

The performance of the recognizer can be enhanced by passing to it more information about salient features in the word. A number of useful features can be easily discerned from the handing out that has previously been performed on the writing: Horizontal histogram, vertical histogram, radial, input and output histograms. Histograms are calculated from each character is given

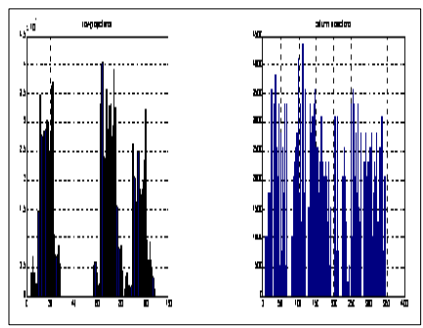

Fig.7. Horizontal histogram and Vertical histogram,

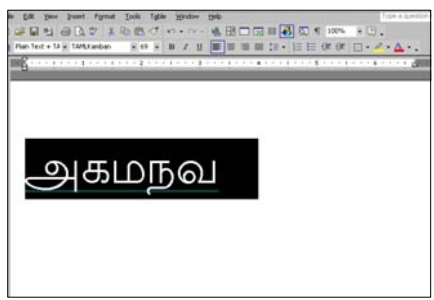

Fig.8. – Final word document of recognized output of given image

## VI. CONCLUSION

From these image processing algorithms and predefined pattern helps to recognize the Tamil characters. So we concluded that these kind of image processing plays important role in both theoretical and practical way of doing. We are trying to identify all the Tamil characters through the complex image processing algorithms and increasing the recognition patterns of the character.

## REFERENCES

[1] R. M. Bozinovic and S. N. Srihari, "Off-line cursive script word recognition", IEEE Trans. On Pattern Anal. Mach. Intell., vol. 11, no. 1, pp. 68-83, Jan. 1989.
[2] H. Bunke, P.S.P. Wang, "Handbook of Character Recognition and Document Image Analysis", World Scientific, 1997.
[3] Abhijit, S. P. and Macy, R. B., Pattern Recognition With Neural Networks in C++, CRC Press, 1995
[4] E.Kavallieratou N.Fakotakis G.Kokkinakis "Handwritten characteristics based on the structural characteristics", In proc. ICPR.2000.,pp 634
[5] Jagadeesh Kannan R and Prabhakar R, "An improved Handwritten Tamil Character Recognition
[6] System using Octal Graph", Int. J. of Computer Science, ISSN 1549-3636, Vol 4 (7): 509-516, 2008
[7] Jagadeesh Kumar R, Prabhakar R and Suresh R.M, "Off-line Cursive Handwritten Tamil Characters Recognition", International Conference on Security Technology, page(s): 159 – 164, 2008
[8] Paulpandian T and Ganapathy V, "Translation and scale Invariant Recognition of Handwritten Tamil characters using Hierarchical Neural Networks", Circuits and Systems, IEEE Int. Sym. , vol.4, 2439 –2441, 1993
[9] Ramanathan R, Ponmathavan S, Thaneshwaran L, Arun.S.Nair, and Valliappan N, "Tamil font Recognition Using Gabor and Support vector machines", International Conference on Advances in Computing, Control, & Telecommunication Technologies, page(s): 613 – 615, 2009
[10] Sarveswaran K and Ratnaweera , "An Adaptive Technique for Handwritten Tamil Character Recognition", International Conference on Intelligent and Advanced Systems, page(s): 151 – 156, 2007